\pdfoutput=1

\documentclass[11pt]{article}

\usepackage[preprint]{acl}
\usepackage{subcaption}
\usepackage{times}
\usepackage{latexsym}
\usepackage{float}
\usepackage{graphicx} 
\usepackage{amsmath, mathtools}
\usepackage{float}
\usepackage{booktabs}
\usepackage{graphicx}
\usepackage[most]{tcolorbox}

\usepackage{multirow} 
\usepackage[T1]{fontenc}

\usepackage[utf8]{inputenc}
\usepackage{amsmath}

\usepackage{microtype}
\usepackage{colortbl}
\usepackage{inconsolata}

\usepackage{graphicx}
\newcommand{\zw}[1]{{\color{black}{#1}}}

\title{Enhancing Test-Time Scaling of Large Language Models with \\   Hierarchical Retrieval-Augmented MCTS 
}

\author{
Alex ZH Dou\textsuperscript{2,*},
Zhongwei Wan\textsuperscript{1,*},
Dongfei Cui\textsuperscript{3},
Xin Wang\textsuperscript{1},
Jing Xiong\textsuperscript{4}, \\
\textbf{Haokun Lin}\textsuperscript{5},
\textbf{Chaofan Tao}\textsuperscript{4},
\textbf{Shen Yan}\textsuperscript{6},
\textbf{Mi Zhang}\textsuperscript{1},
\\
\textsuperscript{1}The Ohio State University \quad
\textsuperscript{2}Case Western Reserve University \quad
\textsuperscript{3} Duke University
\\
\textsuperscript{4}University of Hong Kong \quad
\textsuperscript{5}City University of Hong Kong \quad
\textsuperscript{6}ByteDance
\\
\textsuperscript{*}\text{Equal Contribution.}\\
\texttt{Correspondence to: wan.512@osu.edu}\\
\url{https://github.com/SUSTechBruce/R2LLM}
}

\definecolor{greyL}{RGB}{210,233,232}

\begin{document}
\maketitle
\begin{abstract}
Test-time scaling has emerged as a promising paradigm in language modeling, leveraging additional computational resources at inference time to enhance model performance. In this work, we introduce \textbf{R$^{2}$-LLMs}, a novel and versatile hierarchical retrieval-augmented reasoning framework designed to improve test-time scaling in large language models (LLMs) without requiring distillation from more advanced models to obtain chain-of-thought (CoT) training data. \textbf{R$^{2}$-LLMs} enhances inference-time generalization by integrating dual-level retrieval-based in-context learning: (1) At the \textbf{coarse-level}, our approach extracts abstract templates from complex reasoning problems and retrieves similar problem-answer pairs to facilitate high-level in-context learning; (2) At the \textbf{fine-level}, during Monte Carlo Tree Search (MCTS), \textbf{R$^{2}$-LLMs} efficiently retrieves analogous intermediate solution steps from reference mathematical problem datasets, refining step-wise reasoning with the aid of a process reward model (PRM) for scoring. \textbf{R$^{2}$-LLMs} is a robust hierarchical reasoning-augmentation method that enhances in-context-level reasoning while seamlessly integrating with step-level tree search methods. Utilizing PRM, it refines both candidate generation and decision-making for improved reasoning accuracy.  Empirical evaluations on the \textbf{MATH500, GSM8K, and OlympiadBench-TO} datasets achieve relative substantial improvement with an increase up to   
\textbf{16\% } using LLaMA-3.1-8B compared to the baselines, showcasing the effectiveness of our approach in complex reasoning tasks. 
\end{abstract}

\begin{figure*}[ht]
    \centering
    \includegraphics[width=0.95\textwidth]{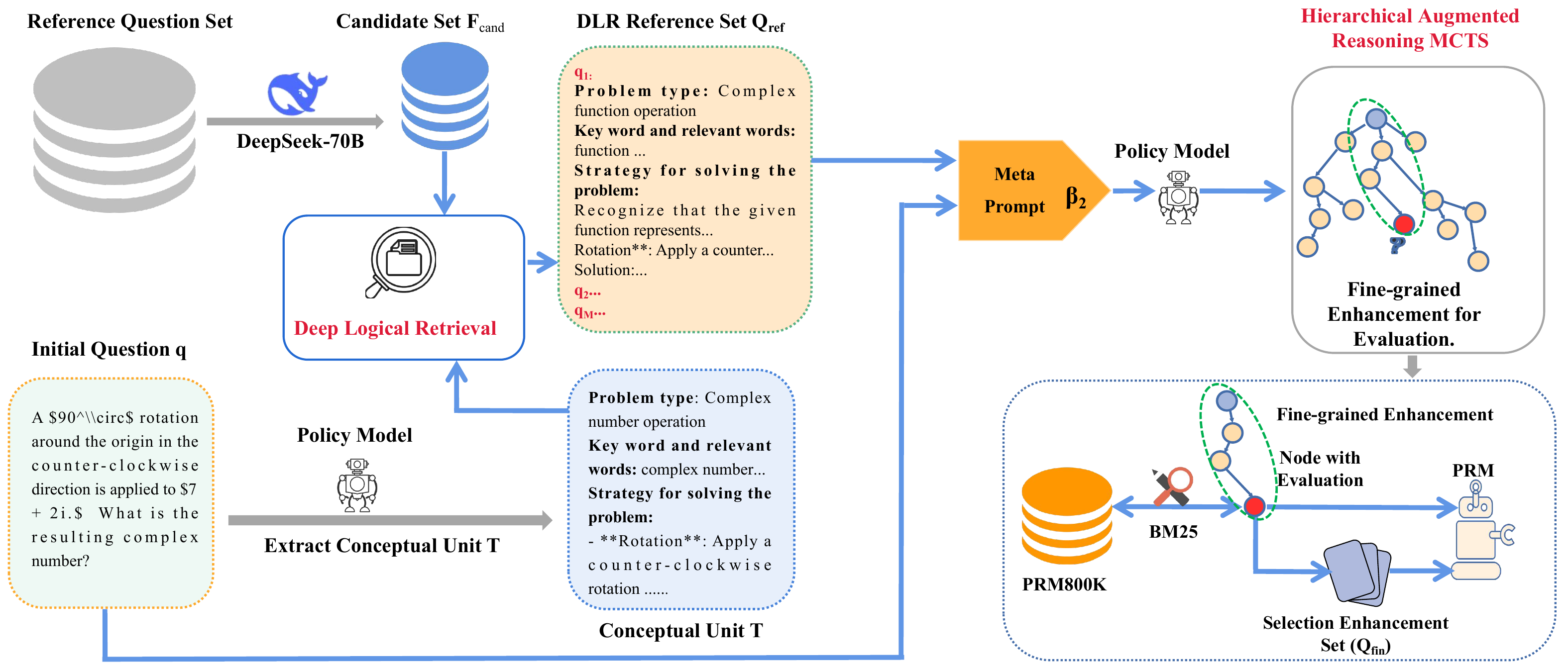}
    \vspace{-0.15in}
    \caption{\small{ 
Illustration of the reasoning process of R$^{2}$-LLMs. R$^{2}$-LLMs employ Hierarchical Augmented Reasoning MCTS to answer the initial question, utilizing two enhancement methods: logical enhancement and fine-grained enhancement.}}
    \vspace{-0.15in}
     \label{fig:enter-label}
\end{figure*}
\section{Introduction}

Emergent abilities of Large Language Models (LLMs)~\cite{wei2022emergent, wan2023efficient, liu2024contemporary, tao2024scaling, wang2024iot} have traditionally relied on increased training-time computation through large-scale generative pretraining~\citep{kaplan2020scaling, hoffmann2022training, wei2022emergent}. Recently, Test-Time Scaling (TTS) has emerged as a complementary paradigm, enhancing reasoning capabilities by allocating extra computational resources at inference~\citep{snell2024scaling}, as validated by DeepSeek-R1~\citep{guo2025deepseek} and OpenAI's O1~\citep{openai2024reasoning}.

Existing TTS approaches are mainly: (1) \textbf{Self-evolution TTS}, which improves reasoning by generating extended Chain-of-Thought (CoT) sequences via large-scale reinforcement learning (RL), exemplified by RLVR~\cite{guo2025deepseek, wan2025srpo, liu2025beyond}; and (2) \textbf{Search-based TTS}, which leverages pre-trained models using inference-time search strategies like Best-of-N~\citep{brown2024large}, Beam search~\citep{snell2024scaling}, and Monte Carlo Tree Search (MCTS)~\citep{zhang2025rest, guan2025rstar}. Search-based methods have gained traction for their efficiency and flexibility, often incorporating Process Reward Models (PRMs) to evaluate intermediate reasoning steps and guide the search effectively\citep{snell2024scaling, wu2024inference, huggingface2024scaling, Wang2023MathShepherdVA}.

Among search-based TTS methods, MCTS demonstrates notable advantages, as mathematical multi-step reasoning tasks inherently involve complex search processes that necessitate systematic exploration of diverse reasoning paths. MCTS excels in managing extensive search spaces by effectively balancing exploration with exploitation, efficiently prioritizing promising candidate paths, and iteratively refining solutions towards optimality~\cite{guan2025rstar}.
However,  conventional PRM+MCTS approaches primarily rely on the information learned during pre-training, which can lead to local optima or exploration blind spots when encountering highly diverse or underrepresented problem distributions~\citep{zhang2025rest}. Moreover, these methods depend solely on the PRM to evaluate steps within MCTS, which may fail to capture global problem-solving strategies and semantic relationships. As a result, the reward signals guiding the search process can be sparse or suboptimal, reducing overall efficiency and accuracy. This limitation increases the risk of deepening the search along incorrect trajectories, ultimately leading to failure in complex reasoning tasks. These challenges underscore the necessity for a more effective and generalizable inference scaling approach—one that enhances reasoning capabilities without requiring extensive additional training while offering a plug-and-play search strategy to improve robustness and adaptability across diverse problem settings.

To enhance the precision of reasoning path exploration, we propose \textbf{R$^{2}$-LLMs} that leverages external retrieval to enhance inference-time generalization through a dual-level retrieval-based in-context learning mechanism.
\textbf{For coarse-level}, we propose Deep Logical Retrieval in section~\ref{inference_reference_set}. Our approach retrieves  analogous problem-answer pairs via abstract problem templates to provide diverse exemplars, enabling the model to capture underlying patterns and variability in problem structures. This facilitates more effective in-context learning, enhancing the model’s adaptability to unseen problems.  \textbf{For fine-level}, we further introduce Hierarchical Augmented Reasoning MCTS in section~\ref{Fine_grained_PRM}, . During MCTS, R$^{2}$-LLMs dynamically retrieves relevant intermediate solution steps from external mathematical problem datasets, enriching the reasoning process with similar prior knowledge. By incorporating these retrieved steps, PRM can provide more informed and contextually consistent evaluations, reducing the risk of inefficient exploration.

Empirical results demonstrate that the proposed retrieval-augmented steps enable R$^{2}$-LLMs to generalize more effectively to complex and unseen problems by leveraging diverse problem-solving strategies from reference datasets. This mitigates the limitations of relying solely on the immediate problem context and significantly enhances reasoning performance. Our approach is evaluated on policy models LLaMA 3.1-8B~\citep{dubey2024llama} and Qwen 2-7B~\citep{yang2024qwen2}, \zw{outperforming ICL-based and tree-based baselines on MATH500~\citep{hendrycks2021measuring}, GSM8K~\citep{cobbe2021training}, and OlympiadBench-TO~\citep{he2024olympiadbench}.}

\vspace{-0.5em}
\section{Related Works}
\vspace{-0.5em}
\paragraph{Test Time Scaling for LLMs.}
Scaling inference-time compute has emerged as a compelling paradigm for enhancing the performance of LLMs ~\citep{openai2024reasoning, guo2025deepseek}. Early work in this area explored techniques such as majority voting ~\citep{wang2022self} and best-of-N methods ~\citep{brown2024large, li2023making}, which generate multiple candidate solutions and select the most frequent or highest-scoring output. More advanced approaches have leveraged search-based strategies, including Monte Carlo Tree Search (MCTS) ~\citep{choi2023kcts, zhang2023planning, liu2024don, zhou2023language}, to systematically explore the reasoning space and improve accuracy. 
To further enhance search efficiency, recent studies have integrated Process Reward Models (PRMs) to guide the selection of high-quality reasoning paths~\citep{setlur2024rewarding, snell2024scaling, lightman2023let, luo2024improve, Wang2023MathShepherdVA}. These models provide refined, step-wise evaluations, particularly beneficial in complex reasoning tasks. Additionally, methods such as BoT ~\citep{yang2024buffer} employ historical thought templates to steer exploration, achieving notable improvements in inference efficiency. ReasonFlux~\citep{Yang2025ReasonFluxHL} adaptively scales fundamental and essential thought templates for simplifying the search space of complex reasoning. 
In contrast, our proposed \text{R$^{2}$-LLMs} framework employs a hierarchical retrieval-augmented strategy that leverages external reference data at both coarse and fine levels, enriching in-context learning and refine intermediate solution steps during MCTS to enhance PRM evaluations.

\paragraph{Mathematical Reasoning.}

%
Mathematical reasoning has long been one of the most challenging tasks in artificial intelligence. Early efforts relied on rule-based methods ~\citep{feigenbaum1963computers, fletcher1985understanding}, but the advent of large language models has shifted the focus toward enhancing reasoning capabilities both during training—via fine-tuning with high-quality mathematical data ~\citep{shao2024deepseekmath, yang2024qwen2, lewkowycz2022solving, yue2023mammoth}—and at inference time through prompt engineering~\citep{wei2022chain} and self-refinement techniques~\citep{madaan2024self, gou2023critic, ke2024critiquellm}.  More recently, researchers have advanced stepwise reasoning by decomposing complex problems into individual reasoning steps. Approaches such as Tree of Thoughts~\citep{yao2023tree} and Monte Carlo Tree Search (MCTS)~\citep{zhang2024llama, chen2024step, feng2023alphazero, zhu2022solving} explore multiple solution paths, with Process Reward Models (PRMs)~\citep{lightman2023let, luo2024improve} providing real-time verification to prune suboptimal paths. While these methods improve accuracy, they often depend on internal model knowledge and struggle with diverse or unseen problems.
In contrast, our proposed R$^{2}$-LLMs framework uses a hierarchical retrieval-augmented approach to boost test-time scaling for mathematical reasoning by integrating external reference data. Unlike previous methods that depend solely on internal reasoning and risk local optima, our approach enriches the process with diverse, contextually relevant examples and intermediate steps.

\paragraph{In-context learning with relevant samples}
In-context learning is a cost-effective guidance approach that enhances model output quality by leveraging similar examples, eliminating the need for fine-tuning \cite{zhou2024mystery,dong2022survey}. Specifically, CoT \cite{wei2022chain,kojima2022large} guides the model’s reasoning process. Self-Consistency (SC) \cite{wang2022self} enhances performance by generating multiple reasoning paths and selecting the most consistent outcome. In addition, Buffer of thought (BoT) \cite{liu2024don} enhances large language model reasoning by utilizing high-level thought templates, shifting the focus beyond problem-level in-context learning. Different from BoT’s template matching, R$^{2}$-LLMs employs a hierarchical retrieval-augmented framework to dynamically integrate global strategies and local reasoning for enhanced problem solving.

\section{Method}
\textbf{Overview of R$^{2}$-LLMs.} In this section, we present a detailed overview of R$^{2}$-LLMs, with the specific process illustrated in Figure \ref{fig:enter-label}. In Section \ref{Preliminary}, we briefly introduce the preliminary of MCTS.
When solving an initial mathematical question \( q \), we \zw{efficiently} extract the \textcolor{black}{\textbf{conceptual unit}}  \( T \) (Section \ref{inference_factor}), which captures the core information of \( q \) and serves as the basis for retrieving a relevant \textcolor{black}{\textbf{DLR reference set}} \( Q_{\text{ref}} \) (Section \ref{inference_reference_set}).
Leveraging \( Q_{\text{ref}} \), we employ MCTS to conduct hierarchical augmented reasoning MCTS  for \( q \) (Section \ref{Fine_grained_PRM}). 


\subsection{MCTS Preliminary}
\label{Preliminary}

MCTS is a heuristic search algorithm that incrementally builds a tree using stochastic simulations. Unlike Minimax, it selects actions via statistical sampling and refines estimates with more simulations. In this paper, we define the MCTS as $MCTS(\cdot)$.
The MCTS algorithm consists of four main phases:

\noindent \textbf{Selection.}  
Starting from the root node, the algorithm recursively selects child nodes until it reaches a leaf node. A policy guides the selection process, often the Upper Confidence Bound for Trees (UCT), which balances exploration (trying less-visited nodes) and exploitation (favoring nodes with higher rewards).

\begin{equation}
\small
\text{UCT}(v) = \frac{Q(v)}{N(v)} + c \sqrt{\frac{\ln N(\text{parent}(v))}{N(v)}}, 
\end{equation}
where $Q(v)$ is the total reward of node $v$, $N(v)$ is the visit count of node $v$, and $c$ is a constant controlling the balance between exploration and exploitation.

\noindent \textbf{Expansion.}  
If the selected leaf node is not terminal, the algorithm expands it by adding child nodes for possible actions, progressively exploring new parts of the search space.

\noindent \textbf{Simulation (Rollout).}  
A simulation starts from the expanded node, taking random or policy-driven actions until reaching a terminal state. This \textit{Rollout} estimates the node's reward.

\noindent \textbf{Backpropagation.}  
The simulation results update node statistics (e.g., total reward, visit count) from the expanded node to the root, refining node quality estimates iteratively.

\subsection{Conceptual Unit Extraction}
\label{inference_factor}

Some studies \cite{zhang2025booststep,yang2024buffer} indicate that highly relevant questions, along with their reasoning steps or solution templates related to the initial question, can enhance policy models' reasoning abilities and improve their problem-solving accuracy.
However, mathematical questions often involve various types, logical conditions, and constraints, forming intricate logical structures. 
%
Relying solely on surface-level semantic information makes it challenging to directly determine the correlation between different problems. 
Therefore, it is essential to extract generalization representations from these questions, allowing for effective categorization and the identification of connections across different questions types. Doing so promotes deeper analysis and a more comprehensive understanding.

Inspired from previous works \cite{yang2024buffer,wu2024beyond}, we extract the generalization features of the initial question from three key perspectives: \textbf{problem types}, \textbf{key terms}, and \textbf{relevant solution strategies}. We collectively refer to this triplet as the \zw{conceptual unit}, denoted by $
T = \left( t_{\text{type}}, t_{\text{key}}, t_{\text{strategy}} \right),
$ 
where \( t_{\text{type}} \) denotes the problem type, \( t_{\text{key}} \) represents the key terms within the problem, and \( t_{\text{strategy}} \) corresponds to the associated solution strategy. The inference factor $\text{T}$ can be obtained as:
\begin{equation}
\small
    T = LLM(\beta_1(x)),
    \label{extra}
\end{equation}
where \( \beta_1(\cdot) \) denotes the meta prompt used for extracting the \zw{conceptual unit} and $x$ is original question. 
The detailed extraction process is provided in the Appendix \ref{meta}.

\begin{table*}[h]
\centering
\small
\vspace{-0.15in}
\caption{Comparative performance of reasoning methods across three benchmark datasets. The best results in each box are highlighted in \text{bold} for clarity.}
\begin{tabular}{c|c|ccc|c}
\midrule[1.2pt]
\multirow{2}{*}{Model}                 & \multirow{2}{*}{Method} & \multicolumn{3}{c|}{Dataset}                  & \multirow{2}{*}{Average} \\ \cmidrule(lr){3-5}
                                       &                          & 
  \textbf{MATH500}          & \textbf{GSM8K}         & \textbf{OlympiadBench}   &                          \\ \midrule
\multirow{4}{*}{LLaMA-3.1-8B-Instruct} & Zero-shot CoT~\citep{kojima2022large}           & 18.0          & 61.5          & 15.4          & 31.6                     \\
                                       & Few-shot CoT~\citep{wei2022chain}             & 47.2          & 76.6          & 16.3          & 46.7                     \\
                                       & CoT+SC@4~\citep{wang2022self}                & 44.2          & 80.5          & 16.5          & 47.1                     \\
                                       & \cellcolor{greyL}\textbf{R$^{2}$-LLMs}                     & \cellcolor{greyL}\textbf{52.5} & \cellcolor{greyL}\textbf{87.4} & \cellcolor{greyL}\textbf{23.7} & \cellcolor{greyL}\textbf{54.5}            \\ \midrule
\multirow{4}{*}{Qwen2-7B-instruct}     & Zero-shot CoT~\citep{kojima2022large}             & 36.9          & 76.6          & 21.3          & 44.9                     \\
                                       & Few-shot CoT~\citep{wei2022chain}             & 52.9          & 85.7          & 21.6          & 53.4                     \\
                                       & CoT+SC@4~\citep{wang2022self}                 & 55.6          & 87.7          & 21.7          & 55.0                     \\
                                       & \cellcolor{greyL} \textbf{R$^{2}$-LLMs}                      & \cellcolor{greyL}\textbf{60.6} & \cellcolor{greyL}\textbf{89.1} &  \cellcolor{greyL}\textbf{28.5} & \cellcolor{greyL}\textbf{59.4}            \\ \midrule[1.2pt]
\end{tabular}
\vspace{-0.15in}
\label{main_result}
\end{table*}

\subsection{Deep Logical Retrieval}
\label{inference_reference_set}

In this section, we propose deep logical retrieval (DLR) to assist the policy model in effective reasoning. Given an initial question \( q \) and its conceptual unit \( T \), DLR aims to retrieve several questions with similar inference logic along with their corresponding reasoning steps to serve as references for the policy model. These similar questions and reasoning steps help the model better understand the reasoning path, thereby enhancing its reasoning capability and efficiency.


Given a set of reference questions, we use DeepSeek-70B \cite{guo2025deepseek} to generate a conceptual unit using Eq.~\ref{extra} and further construct the candidate set $F_{cand} = \{(q_i, T_i, s_i)\}_{i=1}^n$, where $s_i$ represents the solution steps for the $i$-th question.
For a candidate set $F_{cand}$ consisting of $n$ questions, solution steps and their corresponding conceptual units $T_i$, we implemented a two-stage selection process—\text{Preliminary Filtering} followed by \text{Refined Selection}—to identify questions that exhibit deep logical relevance to the initial question $q$. 
We describe this two-stage retrieval process as follows:



\paragraph{Preliminary Filtering.} In the preliminary filtering stage, we employ the BM25~\citep{robertson1976relevance} algorithm to retrieve the most relevant questions from the candidate set \( F_{cand} \), based on the given query \( q \) and its corresponding problem type \( t_{\text{type}} \).
Specifically, we construct a query pair \( (q, t_{\text{type}}) \) of initial questions and compute its similarity with subset of \zw{conceptual unit} from the candidate set  to construct \( \{(q_i, t_{\text{type}})_i\}_{i=1}^{n} \) using BM25. 
The candidate questions are then ranked by their BM25 scores, and the top \( N \) most relevant ones are selected as the coarse-level selection set $Q_{\text{ref}}^{coa}$. The coarse-level set is constructed by filtering candidate questions based on the semantic similarity of the query \( q \) and its corresponding problem type \( t_{\text{type}} \). The selected questions not only belong to the same category as the initial question but also share a similar knowledge foundation in the problem-solving process, ensuring greater accuracy and relevance for subsequent matching.

\paragraph{Refined Selection.} After obtaining coarse-level set \( Q_{\text{ref}}^{coa} \), we further perform fine-grained selection among the these questions. 
Specifically, we \zw{utilize} \( (t_{\text{key}}, t_{\text{strategy}}) \) from the \zw{conceptual unit $T$} of the initial question \( q \) and compare them with the set \( \{(t_{\text{key}_i}, t_{\text{strategy}_i})\}_{i=1}^{N} \) $\in$ \( Q_{\text{ref}}^{\text{coa}} \) to compute the semantic similarity score as:
\begin{equation}
\small
    \begin{aligned}
        e_i &= E\left(t_{\text{key}_i}, t_{\text{strategy}_i}\right), 
        e = E\left(t_{\text{key}}, t_{\text{strategy}})\right) \\
    \end{aligned}
\end{equation}
\begin{equation}
\small
    S_{\text{ref}, i} = Cosine(e_i, e),
\end{equation}
where $E(\cdot)$ is an LM encoder, specifically a pre-trained SentenceBERT \cite{reimers2019sentence}.
$S_{\text{ref}, i}$ is used to measure the cosine similarity between the encoded representations of question keywords and question-solving strategies. By integrating the model's predicted solving strategies and the keywords extracted from the problem, it further assesses the relationship between the initial question \( q \) and identifies candidate questions \( q_i \in Q^{coa}_{\text{ref}} \) that share similar problem-solving knowledge and reasoning approaches.
Subsequently, we selected the $M$ most relevant questions and compiled their corresponding solution processes and strategy into the \zw{DLR}  reference set $Q_{\text{ref}}=\{(q_i, s_i,t_{strategy_i}) \mid i \in \underset{i}{\arg\max_M} S_{\text{ref}, i}, i\in [1,N] \}$.


\subsection{Hierarchical augmented reasoning MCTS }
\label{Fine_grained_PRM}

After obtaining the \zw{DLR reference} set \( Q_{\text{ref}} \) for the initial question \( q \), we designed a hierarchical augmented reasoning MCTS approach to solve the problem. This method divides the reasoning process into two main components: \text{Logical Reasoning Enhancement} and \text{Fine-grained Enhancement}. Logical Reasoning Enhancement is tasked with refining the generation of high-quality reasoning steps, whereas Fine-grained Enhancement aims to deliver more accurate evaluations for every node within the MCTS, thereby boosting the precision and efficiency of the decision-making process as a whole.
Next, we will delve into a comprehensive explanation of these two enhancement approaches.

\paragraph{Logical Reasoning Enhancement.}
In Logical Reasoning Enhancement, we utilize a logic-driven guidance mechanism to enable MCTS to produce high-quality and coherent solution paths.
Specifically, during the MCTS reasoning process, we utilize $Q_{\text{ref}}$ as a reference to steer the policy model $P(\cdot)$ in producing the subsequent node $v_i$ at $i$-th state, leveraging the preceding set of node states $\text{V}_{i-1} = \{v_1, \ldots, v_{i-1}\}$ using meta prompt $\beta_2(\cdot)$:
\begin{equation}
\small
    v_i = P\left(\beta_2\left((q,Q_{\text{ref}}),\text{V}_{i-1}\right)\right).
\end{equation}
Logical Reasoning Enhancement empowers the policy model to draw insights from logically analogous problems, thereby enhancing its ability to generate high-quality candidate solutions. By leveraging established logical patterns and structures, this approach guides the model in delivering more precise and contextually relevant answers.

\paragraph{Fine-grained Enhancement.}
At the \( i \)-th state, the policy model generates \( U \) candidate nodes \( \text{V}_{i}^{cand} = \{v_{i,j}\}_{j=1}^{U} \) based on the previous state node \( V_{i-1} \). Among these candidates, the node with the highest \( Q \) value is selected as the final node for the current state, i.e., $
v_i = \underset{v\in V_{i}^{cand}}{\arg\max} Q(v)$.
It ensures that at each step, the most optimal successor node is chosen to efficiently construct the path.

Aligned with previous research \cite{zhang2025rest}, we use PRM to approximate each node values \( Q(v_{i,j}) \), where \( v_{i,j} \in \text{V}_i^{cand}\).
To further enhance the accuracy of PRM’s value estimation,
we propose a fine-grained (FG) enhancement evaluation approach. We begin by using \( \hat{\text{V}}_{i,j} =  (q, v_1, ..., v_i, v_{i,j}) \), \zw{where $ v_1, ..., v_i$ represent previous steps,} as a query to perform BM25 retrieval within a fine-grained set $F_{FG}$, retrieving a selection enhancement set \( Q_{\text{fin}_{i,j}} \) \zw{with size $K$} that contains questions and reasoning steps relevant to \( q \). 
Each reasoning step is assigned a relevance score $R(v_{i,j})$, which aids in evaluating the values of the node \( v_{i,j} \):
\begin{equation}
\small
    R(v_{i,j}) = R_{\text{PRM}}\left(\beta_3(\hat{\text{V}}_{i,j}, Q_{\text{fin}_{i,j}})\right),
\end{equation}
where \( R_{\text{PRM}}(\cdot) \)  denotes the evaluation score generated by PRM, and \( \beta_3(\cdot) \) represents the meta prompt that assists PRM in enhancing the scoring process.

\section{Experiment}

\subsection{Experiment setting}

\textbf{Policy and Reward Models.} We use three LLMs as policy models: LLaMA 3-8B, LLaMA 3.1-8B \cite{dubey2024llama}, and Qwen 2-7B \cite{yang2024qwen2}. For PRM, we adopt Mistral-7B \cite{tang2024mathscale}, trained on PRM800K\footnote{The Mistral-7B PRM model is open-source: https://huggingface.co/peiyi9979/math-shepherd-mistral-7b-prm}. Notably, we use a logit-based PRM approach rather than step-wise prompting.

\noindent \textbf{Evaluation Benchmark.} We test our method on three challenging open source mathematical benchmarks: MATH500 \cite{hendrycks2021measuring}, focused on high school-level competition mathematics; GSM8K \cite{cobbe2021training}, covering middle school to early high school level problems; and OlympiadBench-TO \cite{he2024olympiadbench}, designed for problems at the level of international mathematics olympiads.


\noindent\textbf{Candidate Set Selection.} For MATH500, we randomly select 2,500 questions and reasoning steps from PRM800K due to its rich and complex mathematical reasoning, which aligns well with their characteristics. For GSM8K, we chose the same number of questions from MAWPS \cite{koncel2016mawps} and MATHQA \cite{amini2019mathqa}, as MAWPS offers various application questions and AQuA includes multiple choice questions based on reasoning, covering GSM8K's real-world math scenarios. For all tests, the candidate set size is 2500. DeepSeek-70B \cite{guo2025deepseek} generated all conceptual units within this set, including the reasoning steps for questions sourced from MAWPS and MATHQA. Regarding OlympiadBench-TO, we choose OpenThoughts \footnote{https://huggingface.co/open-thoughts/OpenThinker-7B}.


\noindent\textbf{Numbers of DLR Reference Set $Q_{ref}$ and Selection Enhancement Set $Q_{fin}$.} The DLR reference set maintains consistency in both question selection and candidate set. By default, the DLR reference set consists of 4 samples. Additionally, for the selection enhancement set, we selected samples from PRM800K, with a set size of 3. \zw{We show the sensitively analysis in Section~\ref{sec: Sensitively analysis}.}

\noindent\textbf{Baseline.} We primarily compare our approach against three traditional example-based ICL methods: zero-shot CoT \cite{kojima2022large}, few-shot CoT \cite{wei2022chain}, and SC+CoT \cite{wang2022self}. For SC, we conduct four sampling iterations, referred to as CoT+SC@4. Additionally, we compare our approach with various tree-based structures, including ToT \cite{yao2023tree}, RAP \cite{hao2023reasoning}, ReST-MCTS$^*$ \cite{zhang2025rest} and LiteSearch \cite{wang2024litesearch}.

\noindent\textbf{Evaluation metrics.} We use accuracy (\%) as the evaluation metric, where a solution is correct only if the model's final answer exactly matches the ground truth.. A solution is deemed correct only if the final reasoning process is fully aligned with the ground truth.

In addition, the more experiment can be seen in Appendix \ref{extra_experiment}.

\subsection{Performance on Various Reasoning Benchmarks}

Table \ref{main_result} compares the reasoning performance of LLaMA-3.1-8B-Instruct and Qwen2-7B-instruct across three datasets (MATH, GSM8K, OlympiadBench) using four reasoning methods. R$^{2}$-LLMs achieves the highest scores, with Qwen2-7B-instruct outperforming LLaMA-3.1-8B-instruct in all settings. For example, on MATH, Qwen2-7B-Instruct reaches 60.6\% with the proposed method, compared to LLaMA-3.1-8B-Instruct's 52.5\%. Few-shot CoT and CoT+SC@4 show notable improvements over Zero-shot CoT; for instance, LLaMA-3.1-8B-Instruct's GSM8K score rises from 61.5\% (Zero-shot CoT) to 80.5\% (CoT+SC@4). Meanwhile, our method also shows a significant improvement on OlympiadBench.


\subsection{Comparison with Other Tree-based Methods}

To further assess the effectiveness of R$^{2}$-LLMs, we conducted comparative experiments on the MATH and GSM8K datasets against leading tree-based approaches. Specifically, we selected ToT, RAP, ReST-MCTS$^{*}$, and LiteSearch as baselines and utilized the widely adopted Llama-3-8B-Instruct as the backbone model for inference, ensuring fairness and comparability in our evaluation.

Table \ref{tree_base} compares tree-based methods on GSM8K and MATH. R$^{2}$-LLMs achieve the best results—84.6\% on GSM8K and 34.7\% on MATH—outperforming all baselines. Notably, it surpasses LiteSearch by 2.3\% and RAP by 4.1\% on GSM8K, and leads ReST-MCTS by 3.3\% on MATH, showing strong performance on both arithmetic and complex reasoning tasks. Compared with other methods, R$^{2}$-LLMs outperforms existing approaches by combining coarse-level retrieval for global strategy, which aids in handling rare problems, with fine-grained retrieval that enriches node evaluation using external steps, effectively mitigating sparse rewards in standard MCTS and other tree-based methods.


\begin{table}[t]
\vspace{-0.15in}
\caption{Comparison of tree-based methods on GSM8K and MATH. \textbf{Bold} indicates the best performance.}
\resizebox{0.48\textwidth}{!}{
\begin{tabular}{c|c|cc}
\midrule[1.2pt]
\multirow{2}{*}{Model}                 & \multirow{2}{*}{Method} & \multicolumn{2}{c}{Dataset}   \\ \cmidrule(lr){3-4} 
                                       &                         & \textbf{GSM8K}         & \textbf{MATH500}          \\ \midrule
\multirow{5}{*}{LLaMA-3-8B-Instruct} & ToT                     & 69.0          & 13.6          \\
                                       & RAP                     & 80.5          & 18.8          \\
                                       & ReST-MCTS$^{*}$         & -             & 31.4          \\
                                       & LiteSearch              & 82.3          & -             \\
                                       & \cellcolor{greyL} \textbf{R$^{2}$-LLMs}                    & \cellcolor{greyL}\textbf{84.6} & \cellcolor{greyL}\textbf{34.7} \\ \midrule[1.2pt]
\end{tabular}
}
\label{tree_base}
\vspace{-0.15in}
\end{table}

\begin{table*}[t]
\caption{Performance evaluation of the impact of DLR reference sets $Q_{ref}$ on the reasoning capabilities of R$^{2}$-LLMs, tested on the GSM8K and MATH datasets. \textbf{Bold} indicates the best performance.}
\centering
\small
\begin{tabular}{c|c|cccc}
\midrule[1.2pt]
\multirow{2}{*}{Model}                 & \multirow{2}{*}{Datasets} & \multicolumn{4}{c}{Deep Logical Retrieval (DLR) Reference Set}                                                                             \\ \cmidrule(lr){3-6} 
                                       &                           & \multicolumn{1}{c|}{w/o} & \multicolumn{1}{c|}{In Domain}       & \multicolumn{2}{c}{Out of Domain} \\ \midrule
\multirow{4}{*}{LLaMA-3.1-8B-Instruct} & \multirow{2}{*}{\textbf{GSM8K}}    & \multicolumn{1}{c|}{-}             & \multicolumn{1}{c|}{\textbf{MAWPS} \& \textbf{MATHQA}} & \textbf{PRM800K}          & \textbf{AMC 12}         \\
                                       &                           & \multicolumn{1}{c|}{\cellcolor{greyL}82.9}          & \multicolumn{1}{c|}{\cellcolor{greyL}87.4}            & \cellcolor{greyL}86.2             & \cellcolor{greyL} 83.15          \\ \cmidrule(lr){2-6} 
                                       & \multirow{2}{*}{\textbf{MATH}}     & \multicolumn{1}{c|}{-}             & \multicolumn{1}{c|}{\textbf{PRM800K}}         & \textbf{MATHQA}           & \textbf{AMC12}          \\
                                       &                           & \multicolumn{1}{c|}{\cellcolor{greyL}47.5}          & \multicolumn{1}{c|}{\cellcolor{greyL}52.5}            & \cellcolor{greyL}43.5             & \cellcolor{greyL}49.9           \\ 
                                       \midrule[1.2pt]
\end{tabular}
\label{domain}
\end{table*}

\vspace{-0.1in}
\subsection{Domain Impact on the DLR Reference Set }


When selecting the DLR reference set, we generally ensure it is in-domain with the test set. To assess the model's ability to generalize to out-of-domain scenarios, we evaluate its performance on datasets such as GSM8K and MATH. Furthermore, we apply the DLR reference set across different domains to test their adaptability.
Table \ref{domain} presents the impact of the domain relationship between the DLR reference set and the test questions on accuracy. The experiment is based on the LLaMA-3.1-8B-Instruct model and is conducted on two mathematical problem datasets, GSM8K and MATH. In-domain means that the DLR reference set and the test questions come from the same domain, whereas out-of-domain indicates that they belong to different domains.
The experimental results reveal that the model achieves its best performance on in-domain inference sets, where the domain of the questions aligns closely with the DLR reference set. For instance, on the GSM8K dataset, the model attains a score of 87.4\%, demonstrating strong generalization capabilities within the same domain. However, when evaluated using out-of-domain DLR reference set, where the question domain differs significantly, R$^{2}$-LLMs's performance declines noticeably. For example, on the MATH dataset, the score drops to 43.5\%, indicating a substantial performance gap compared to in-domain tasks.
From the results mentioned above, it can be concluded that although the performance degrades across different domains, in most cases, it still helps the model to enhance the overall results.
\subsection{Ablation Analysis}

\begin{table}[h]
\vspace{-1em}
\caption{Ablation analysis of R$^{2}$-LLMs. The best results in each box are highlighted in \textbf{bold} for clarity. DLR denotes Deep Logical Retrieval and FG denotes Fine-grained Enhancement.}
\footnotesize
\resizebox{0.48\textwidth}{!}{  
\begin{tabular}{c|c|cc}
\midrule[1.2pt]
\multirow{2}{*}{Model}                 & \multirow{2}{*}{Method} & \multicolumn{2}{c}{Dataset}   \\ \cmidrule(lr){3-4} 
                                       &                          & \textbf{MATH}          & \textbf{GSM8K}         \\ \midrule
\multirow{4}{*}{LLaMA-3.1-8B-Instruct} & MCTS                     & 46.6          & 82.5          \\
                                       & MCTS$_\text{w/DLR}$         & 50.3          & 86.6          \\
                                       & MCTS$_\text{w/FG}$         & 47.5          & 82.9          \\
                                       & \cellcolor{greyL}MCTS$_\text{w/DLR+FG}$      & \cellcolor{greyL}\textbf{52.5} & \cellcolor{greyL}\textbf{87.4} \\ \midrule
\multirow{4}{*}{Qwen2-7B-instruct}     & MCTS                     & 53.7          & 85.9          \\
                                       & MCTS$_\text{w/DLR}$         & 58.2          & 88.7          \\
                                       & MCTS$_\text{w/FG}$         & 55.6          & 84.8          \\
                                       & \cellcolor{greyL}MCTS$_\text{w/DLR+FG}$      & \cellcolor{greyL}\textbf{60.6} & \cellcolor{greyL}\textbf{89.1} \\ \midrule[1.2pt]
\end{tabular}
} 
\label{abaltion}
\end{table}

To assess the impact of each component on the performance of R$^{2}$-LLMs, we conducted a series of ablation experiments. The baseline MCTS method is compared against three variants: MCTS$_\text{w/DLR}$, which incorporates logical reasoning enhancements, MCTS$_\text{w/FG}$, which integrates fine-grained enhancement, and MCTS$_\text{w/DLR+FG}$ is equal to R$^{2}$-LLMs, which combines both improvements.

Table \ref{abaltion} presents the results of an ablation study evaluating the impact of different components in R$^{2}$-LLMs on the MATH and GSM8K datasets. We conduct experiments using two instruction-tuned language models, LLaMA-3.1-8B-Instruct and Qwen2-7B-Instruct, under different approaches. The results demonstrate that each individual enhancement contributes to performance gains, with the combination of both (MCTS$_\text{w/DLR+FG}$) yielding the best results across both datasets. Specifically, LLaMA-3.1-8B-Instruct achieves 52.5\% on MATH and 87.4\% on GSM8K, while Qwen2-7B-Instruct reaches 60.6\% and 89.1\%, respectively.
In addition, for LLaMA-3.1-8B-Instruct, incorporating logical reasoning enhancements (MCTS$_\text{w/DLR}$) leads to an absolute gain of +3.7\% on MATH (46.6\% → 50.3\%) and +4.1\% on GSM8K (82.5\% → 86.6\%), while adding fine-grained enhancement (MCTS$_\text{w/FG}$) provides a smaller improvement of +0.9\% on MATH (46.6\% → 47.5\%) and +0.4 on GSM8K (82.5\% → 82.9\%). When both components are combined (MCTS$_\text{w/DLR+FG}$), the performance further increases to 52.5\% (+5.9\%) on MATH and 87.4\% (+4.9\%) on GSM8K, demonstrating a synergistic effect.
These findings highlight the effectiveness of our proposed method in improving mathematical reasoning performance.

\subsection{Sensitively Analysis}

\label{sec: Sensitively analysis}

\begin{figure*}[ht]
  \centering
  \begin{minipage}[b]{0.32\linewidth}
    \centering
    \subfloat[Candidate Set ($F_{cand}$) Size $N$\label{gm_can}]{\includegraphics[width=\linewidth]{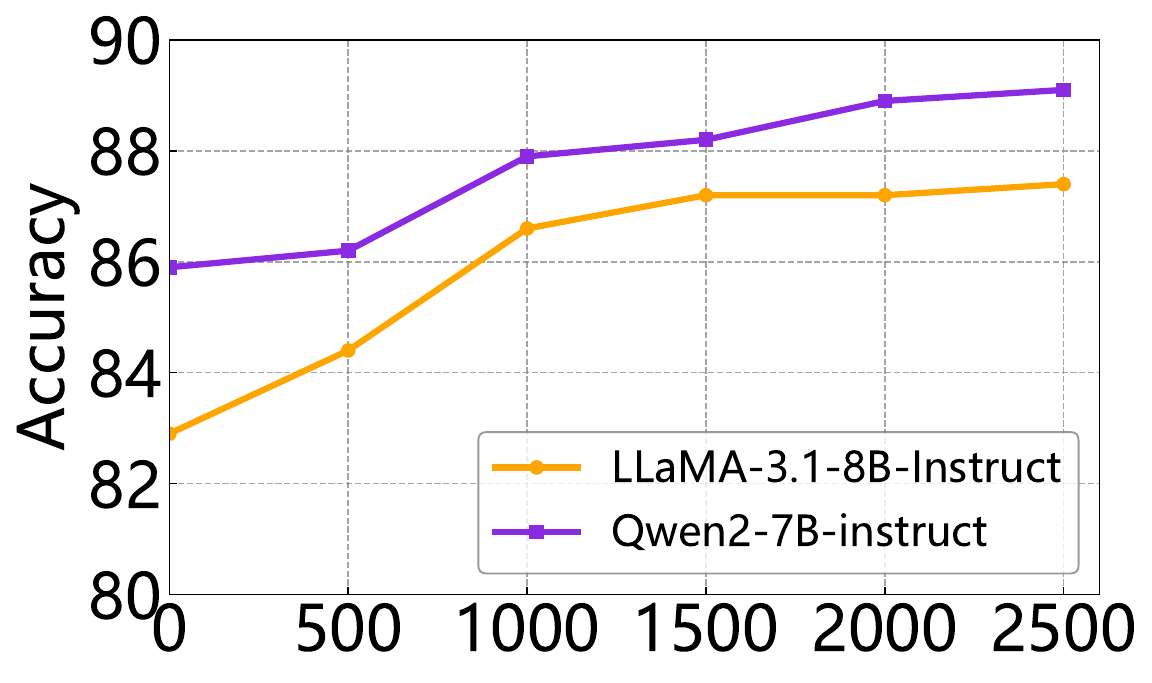}}
  \end{minipage}
  \hfill
  \begin{minipage}[b]{0.32\linewidth}
    \centering
    \subfloat[DLR Reference Set ($Q_{ref}$) Size $M$\label{gm_icl}]{\includegraphics[width=\linewidth]{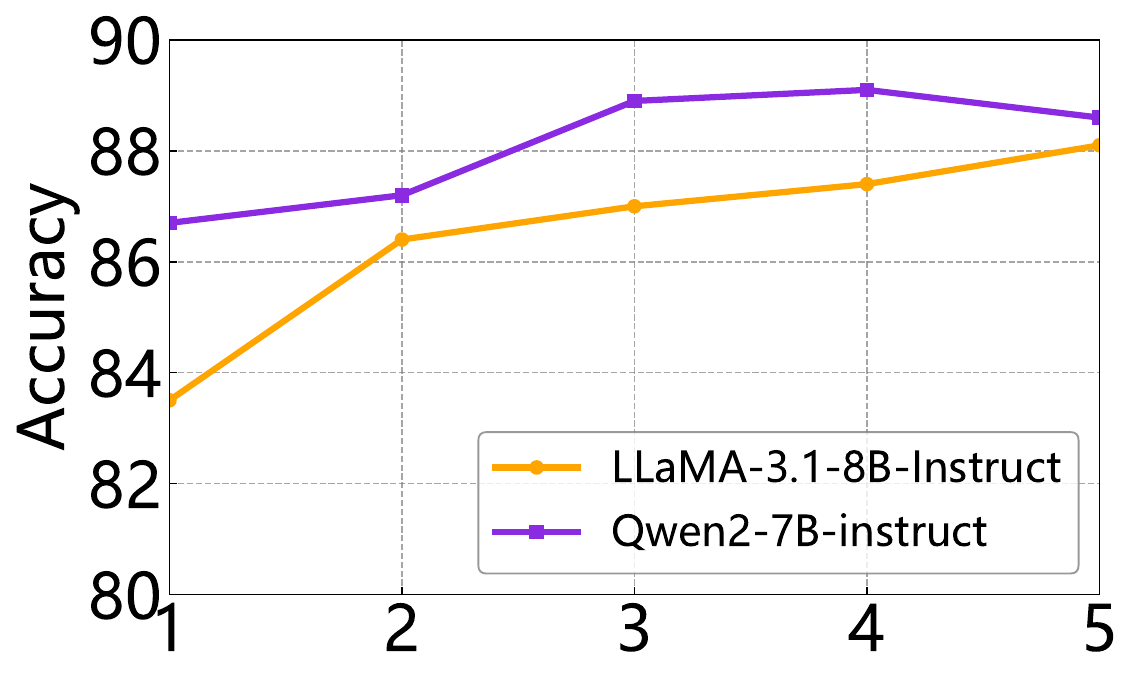}}
  \end{minipage}
  \hfill
  \begin{minipage}[b]{0.32\linewidth}
    \centering
    \subfloat[Selection Enhancement Set ($Q_{fin}$) Size $K$\label{gm_fine}]{\includegraphics[width=\linewidth]{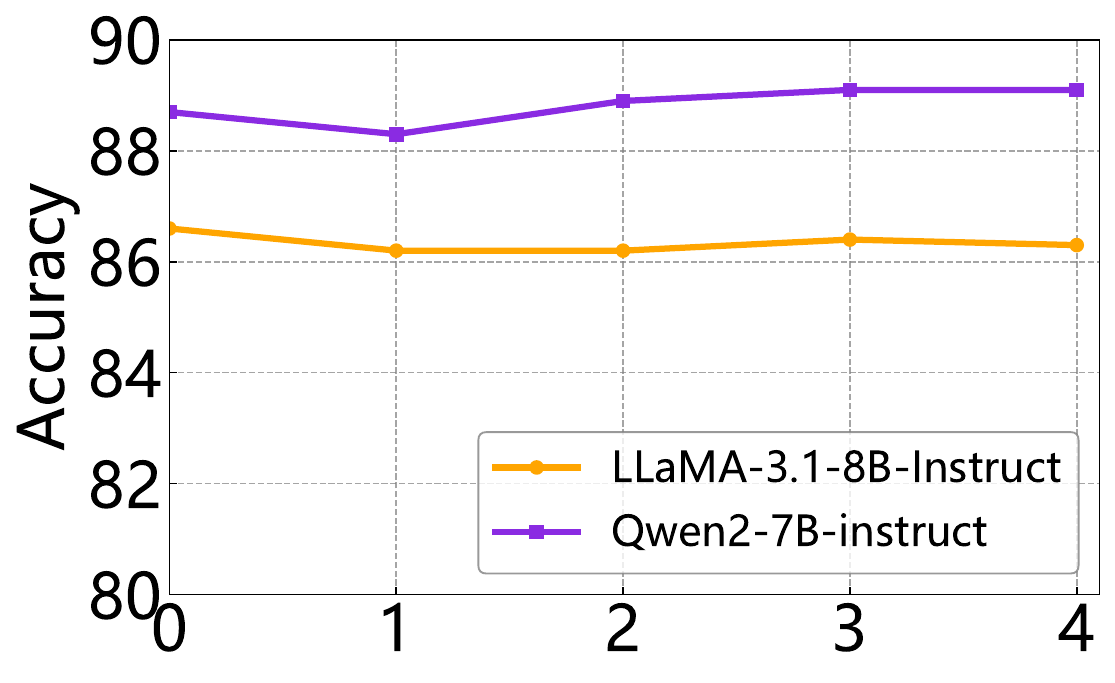}}
  \end{minipage}
  
  \vspace{-1em}
  
  \begin{minipage}[b]{0.32\linewidth}
    \centering
    \subfloat[Candidate Set ($F_{cand}$) Size $N$\label{math_can}]{\includegraphics[width=\linewidth]{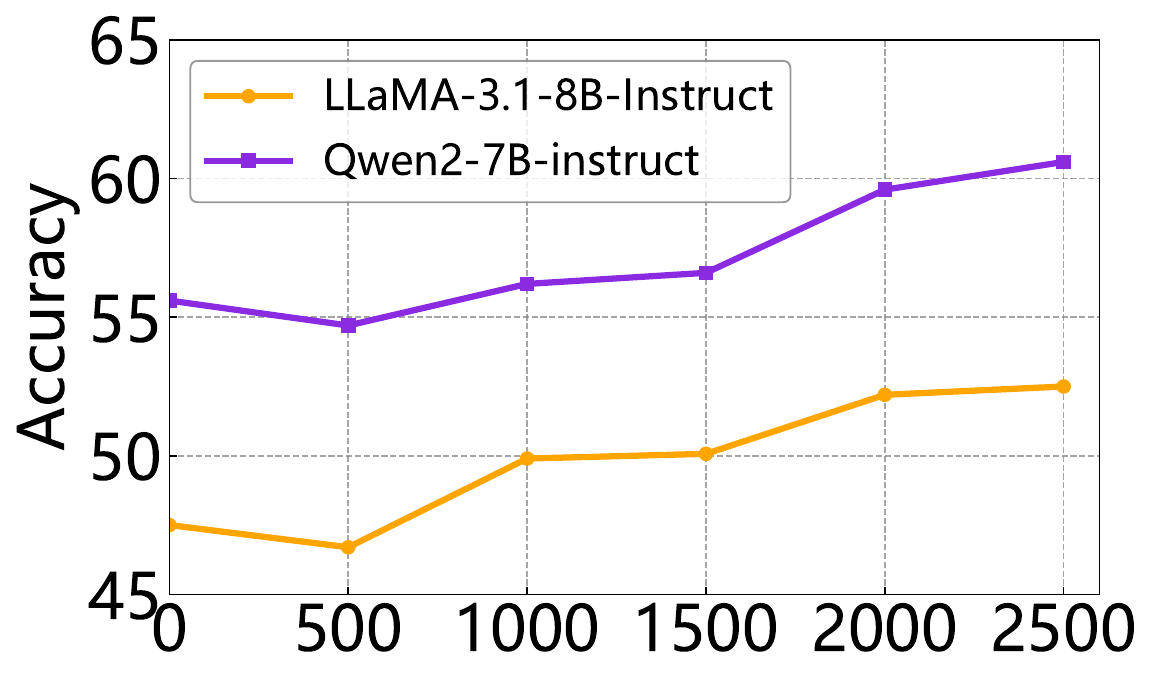}}
  \end{minipage}
  \hfill
  \begin{minipage}[b]{0.32\linewidth}
    \centering
    \subfloat[DLR Reference Set ($Q_{ref}$) Size $M$\label{math_icl}]{\includegraphics[width=\linewidth]{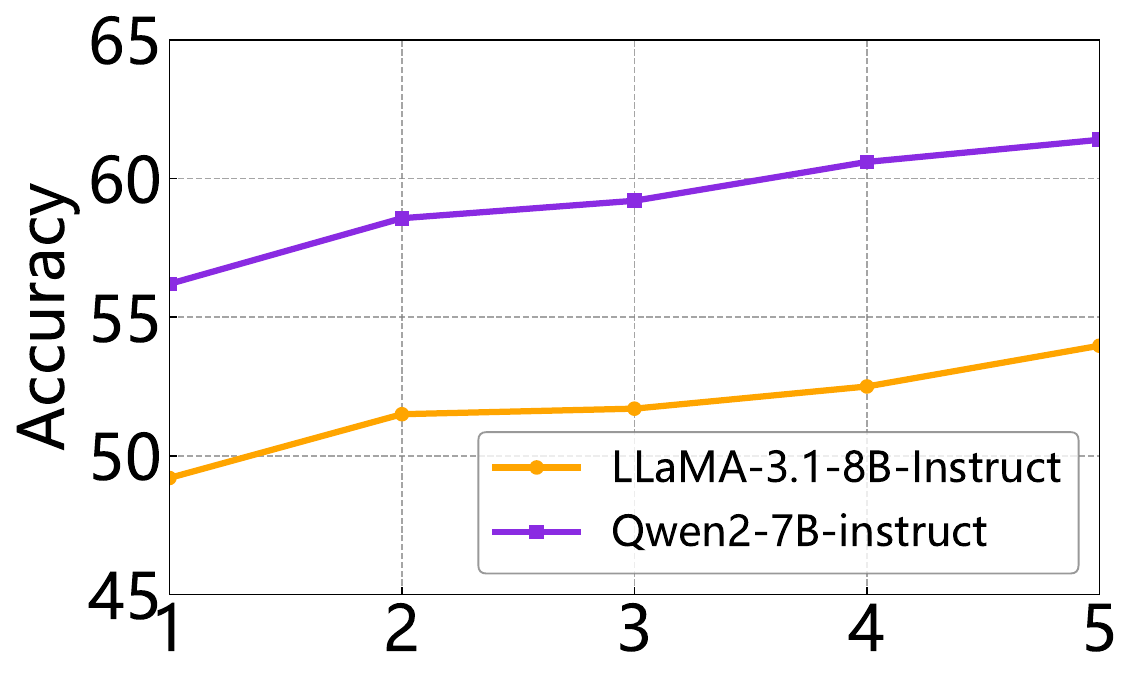}}
  \end{minipage}
  \hfill
  \begin{minipage}[b]{0.32\linewidth}
    \centering
    \subfloat[Selection Enhancement Set ($Q_{fin}$) Size $K$\label{math_fine}]{\includegraphics[width=\linewidth]{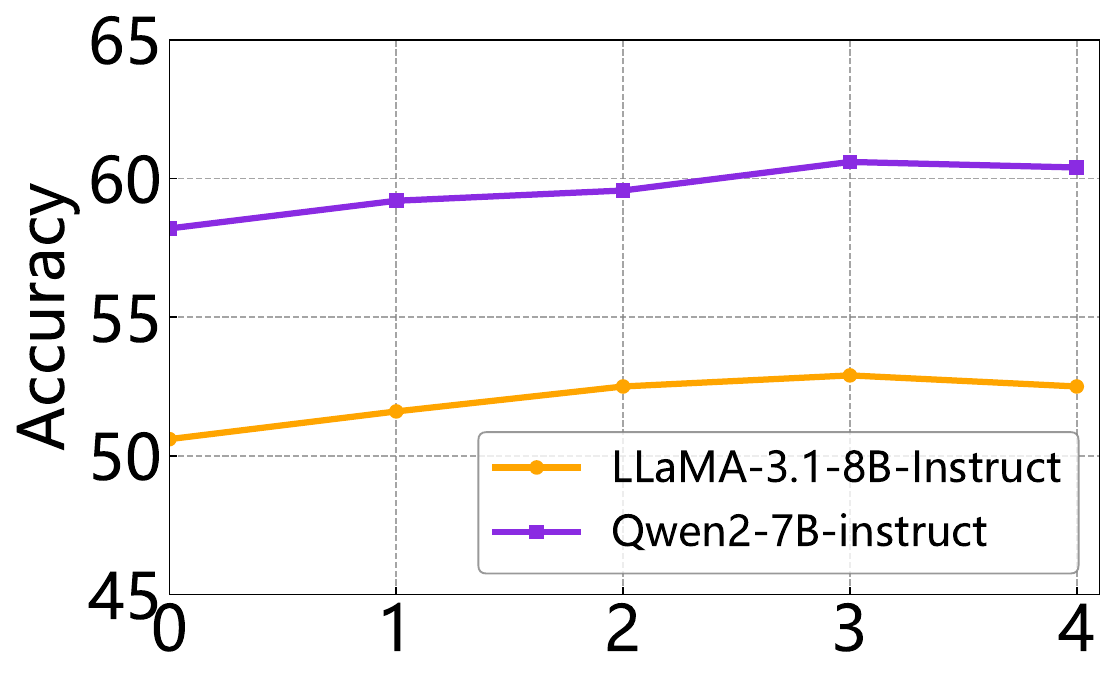}}
  \end{minipage}
  
  \caption{Sensitivity Analysis on GSM8K (top row) and MATH500 (bottom row).}
  \label{sens_STEP_combined}
\end{figure*}

In this section, we examine how the sample size of the candidate set \( F_{cand} \), the DLR reference set \( Q_{\text{ref}} \), and the selection enhancement set \( Q_{fin} \) affects the results, using Qwen2-7B-Instruct and LLaMA-3.1-8B-Instruct as policy models on the GSM8K and MATH datasets. 

\noindent \textbf{Impact of Candidate Set Size.} Figure \ref{gm_can} and Figure \ref{math_can} illustrate the impact of candidate set size on accuracy for the GSM8K and MATH datasets, respectively. Both figures reveal a positive correlation between candidate set size and accuracy. Notably, accuracy increases sharply as the sample size grows from 1000 to 1500, but after reaching 2000, the overall improvement plateaus.

\noindent \textbf{Impact of DLR Reference Set Size.} Figure \ref{gm_icl} and Figure \ref{math_icl} depict the effect of DLR reference set size on accuracy for the GSM8K and MATH datasets. Like the candidate set size, a larger DLR reference set leads to a noticeable improvement in accuracy. When the size reaches 4, accuracy increases by 5\% on MATH and 4.5\% on GSM8K with LLaMA-3.1-8B-Instruct. This indicates that expanding the DLR reference set size can effectively improve the reasoning quality of MCTS, thereby enhancing accuracy.

\noindent \textbf{Impact of Selection Enhancement Set Size.} Figure \ref{gm_fine} and Figure \ref{math_fine} present the influence of selection enhancement set size on the GSM8K and MATH datasets, respectively. While there is still a positive correlation between set size and accuracy, its impact is less pronounced compared to the effect of candidate set size. Specifically, for the MATH dataset, increasing the size to 3 results in only a modest 2.4\% improvement in accuracy.

\section{Conclusion}
In this work, we presented R$^{2}$-LLMs, a hierarchical retrieval-augmented framework that enhances test-time scaling for LLMs by leveraging both Deep Logical Retrieval at the coarse level and Hierarchical Augmented Reasoning MCTS at the fine level. Our approach integrates external reference data to enrich in-context learning and employs a process reward model to refine candidate generation and decision-making. Empirical results, with improvements up to 16\% on key benchmarks, validate the effectiveness of R$^{2}$-LLMs in tackling complex mathematical reasoning tasks.

\section{Limitation}
Our current evaluations have focused primarily on mathematical reasoning benchmarks, leaving its effectiveness in other domains—such as general knowledge, symbolic logic, and multimodal tasks—less explored. Besides, most experiments have been conducted using relatively modest models, and further investigation is needed to understand the performance and scalability of R$^{2}$-LLMs on larger, potentially closed-source models.

\section{Potential Risks}
Our work makes clear contributions by enhancing LLMs’ reasoning abilities. It enables more accurate and trustworthy AI support in complex reasoning tasks such as education, scientific analysis, and decision-making. However, improved reasoning capabilities also pose risks—such as producing persuasive yet inaccurate outputs—especially when reasoning chains are poorly guided. Therefore, responsible and transparent use of such enhanced reasoning frameworks is essential to ensure positive societal impact.

\bibliography{main}

\clearpage
\onecolumn
\appendix

\section{Extra Experiments}
\label{extra_experiment}

\subsection{Comparison with other TTS baselines}

To ensure a fair comparison with other TTS-based methods, we select BoT \cite{liu2024don} as the baseline. As shown in Table~\ref{tts}, we evaluate its performance on two popular base models (LLaMA-3.1-Instruct and Qwen-2-Instruct) across the GSM8K and MATH500 datasets, comparing it with the existing TTS-based method BoT. The results demonstrate that \textbf{R$^{2}$-LLMs} consistently outperforms BoT in all settings. Notably, the improvement is particularly significant on the more challenging MATH500 dataset (e.g., from 25.7 to 52.5, or from 34.5 to 60.6), indicating that our method not only generalizes well across different backbones but also significantly boosts the model's performance on complex mathematical reasoning tasks.

\begin{table}[h]
\centering
\caption{Comparison with another TTS-based method (BoT). The best results in each box are highlighted in \textbf{bold} for clarity.}
\begin{tabular}{c|c|cc}
\midrule[1.2pt]
\textbf{Model}                       & Method & GSM8K & MATH500 \\ \midrule
                                     & BoT    & 62.5  & 25.7    \\
\multirow{-2}{*}{LLaMA-3.1-Instruct} & \cellcolor{greyL}\textbf{R$^{2}$-LLMs}  & \cellcolor{greyL}87.4  & \cellcolor{greyL}52.5    \\ \midrule
                                     & BoT    & 80.4  & 34.5    \\
\multirow{-2}{*}{Qwen-2-Instruct}  & \cellcolor{greyL}\textbf{R$^{2}$-LLMs}  & \cellcolor{greyL}89.1  & \cellcolor{greyL}60.6    \\ \midrule[1.2pt]
\end{tabular}
\label{tts}
\end{table}

\subsection{Time consumption analysis}

\begin{table}[h]
\centering
\caption{The computational overhead incurred by \textbf{R$^{2}$-LLMs}.}
\begin{tabular}{c|cc}
\midrule[1.2pt]
Method                                                    & MATH500 & GSM8k \\ \midrule
 Plain MCTS & 7.00h   & 3.20h \\ \midrule
\cellcolor{greyL}\textbf{R$^{2}$-LLMs}                                                      & \cellcolor{greyL}7.35h   & \cellcolor{greyL}3.45h \\ \midrule[1.2pt]
\end{tabular}
\label{time_comparision}
\end{table}

To evaluate the computational efficiency of our method, we compare the runtime of \textbf{R$^{2}$-LLMs} with the baseline Plain MCTS across two benchmark datasets, as shown in Table~\ref{time_comparision}. On the MATH500 dataset, \textbf{R$^{2}$-LLMs} completes the task in 7.35 hours using 4 A100 GPUs, compared to 7.00 hours required by the baseline. Similarly, on GSM8K, the runtime increases modestly from 3.20 to 3.45 hours. These results demonstrate that \textbf{R$^{2}$-LLMs} achieves significant performance gains with only a marginal increase in computational overhead—approximately 5\% on MATH500 and 7.8\% on GSM8K—validating the practicality and scalability of our method for real-world deployment.


\subsection{Performance with different PRM model}

\begin{table}[h]
\centering
\caption{Performance comparision with different PRM models. The best results in each box are highlighted in \textbf{bold} for clarity.}
\begin{tabular}{c|cc}
\midrule[1.2pt]
Method                                                            & MATH500       & GSM8k         \\ \midrule
\cellcolor[HTML]{FFFDFA}{\color[HTML]{333333} MCTS w/ Mistral-7B} & 46.6          & 82.5          \\ \midrule
\cellcolor{greyL}\textbf{R$^{2}$-LLMs} w/ Mistral-7B                                                & \cellcolor{greyL}52.5          & \cellcolor{greyL}87.4          \\ \midrule
MCTS w/Qwen2.5-7B Math PRM                                        & 47.9          & 84.1          \\ \midrule
\cellcolor{greyL}\textbf{R$^{2}$-LLMs} w/Qwen2.5-7B Math PRM                                        & \cellcolor{greyL}\textbf{53.2} & \cellcolor{greyL}\textbf{89.7} \\ \midrule[1.2pt]
\end{tabular}
\label{better_prm}
\end{table}

To further validate the effectiveness of our approach, we compare \textbf{R$^{2}$-LLMs} with different state-of-the-art PRM (Policy Retrieval Module) backbones, as shown in Table~\ref{better_prm}. When using Mistral-7B as the PRM, \textbf{R$^{2}$-LLMs} achieves substantial improvements over the MCTS baseline, with scores rising from 46.6 to 52.5 on MATH500 and from 82.5 to 87.4 on GSM8K. Similarly, when adopting the more advanced Qwen2.5-7B Math PRM, our method further boosts performance, reaching 53.2 on MATH500 and 89.7 on GSM8K. These results confirm that \textbf{R$^{2}$-LLMs} consistently enhances performance across PRM choices, and benefits even more from stronger PRMs, highlighting the flexibility and scalability of our framework. Due to time constraints, we focus on evaluation using LLaMA 3.1 8B for fair and efficient comparison.

\subsection{Ablation analysis on abstrct template}

\begin{table}[h]
\centering
\caption{Ablation analysis on abstract template using MATH500. The best results in each box are highlighted in \textbf{bold} for clarity. }
\begin{tabular}{c|c}
\midrule[1.2pt]
Method                                                          & MATH500 \\ \midrule
\cellcolor[HTML]{FFFDFA}{\color[HTML]{333333} Without anything} & 47.5    \\ \midrule
Only problem types                                              & 48.7    \\ \midrule
Only key terms                                                  & 49.7    \\ \midrule
Only solution strategies                                        & 50.7    \\ \midrule
Problem types + key terms                                       & 49.7    \\ \midrule
Problem types + solution strategies                             & 50.9    \\ \midrule
Key terms + solution strategies                                 & 52.2    \\ \midrule
\cellcolor{greyL}Problem types + solution strategies + key terms (\textbf{R$^{2}$-LLMs})                & \cellcolor{greyL}52.5    \\ \midrule[1.2pt]
\end{tabular}
\label{abstract_template}
\end{table}

To investigate the effectiveness of each component in the abstract template, we conduct an ablation study on the MATH500 dataset using the LLaMA-3.1-8B Instruct model. As shown in Table~\ref{abstract_template}, the abstract template consists of three elements: problem types, key terms, and relevant solution strategies. Removing all components results in the lowest accuracy of 47.5\%. Adding only problem types slightly improves performance to 48.7\%, while including only key terms or only solution strategies leads to further gains of 49.7\% and 50.7\%, respectively. Combining problem types with key terms does not yield additional benefits (49.7\%), but combining problem types with solution strategies improves the score to 50.9\%. Notably, the combination of key terms and solution strategies achieves a stronger result of 52.2\%. The best performance of 52.5\% is obtained when all three components are included, as used in \textbf{R$^{2}$-LLMs}, confirming that each part of the abstract template contributes incrementally to overall reasoning performance.

\subsection{Results on larger policy model}
\begin{table}[h]
\centering
\caption{Performance using larger policy model (Qwen-2 14B) with MATH500 and GSM8K. }
\begin{tabular}{c|cl}
\midrule[1.2pt]
Method     & MATH500 & GSM8K \\ \midrule
Plain MCTS & 88.5    & 58.4  \\ \midrule
\cellcolor{greyL}\textbf{R$^{2}$-LLMs}       & \cellcolor{greyL}91.6    & \cellcolor{greyL}62.6  \\ \midrule[1.2pt]
\end{tabular}
\label{large_model}
\end{table}

To address the suggestion of evaluating our method on a larger language model, we conduct additional experiments using \textbf{Qwen-2 14B} as the policy model. As shown in Table~\ref {large_model}, our method \textbf{R$^{2}$-LLMs} achieves strong performance improvements over the baseline Plain MCTS on both benchmarks. Specifically, on the challenging MATH500 dataset, accuracy increases from 88.5\% to \textbf{91.6\%}, and on GSM8K, from 58.4\% to \textbf{62.6\%}. These results confirm that \textbf{R$^{2}$-LLMs} consistently enhance performance even with larger-scale models, demonstrating its robustness and scalability across model sizes.

\subsection{Future works and direction}

This work primarily focuses on training-free, retrieval-augmented process reward models for test-time scaling, with an emphasis on post-training reasoning enhancement.
Potential future directions include further improving reasoning ability through integration with RL-based reasoning~\cite{wan2025srpo}, adapting to various multimodal and domain-specific application scenarios~\cite{zhu2025, wan2024look, xiong2024autoregressive, zheng2024structure, liu2024benchmarking, gong2024neuroclips, wan2023spatio, luo2024enhancing, liu2024etp, shen2025efficient, wan2025meda, zhang2025enhancing, liu2024can, huang2024evolver, zhu2024dglf, chen2024clinicalbench, liu2025knowledge, wan2023med, liu2024zero, wan2024electrocardiogram, xu2025swingarena}, using information optimized strategies~\cite{xiong2022self, liang2020large, liang2020many}, developing retrieval-augmented reasoning~\cite{li2024uncertaintyrag, wan2022g} for longer texts, incorporating efficient inference algorithms~\cite{wang2024svd, wan2024d2o, liu2024contemporary, xin2024v, shen2024famba, wang2025svd, xiong2024uncomp, xiong2025parallelcomp}, and supporting dynamic and multimodal datasets~\cite{shen2025phyx, chen2025recent} to fully exploit the post-training reasoning capabilities of LLMs.

\section{Example Appendix}
\label{sec:appendix}

\subsection{Example of DLR set samples}

\textbf{Initial questions:} A \(90^\circ\) rotation around the origin in the counter-clockwise direction is applied to \(7 + 2i.\) What is the resulting complex number?

\vspace{2em}
\textbf{Sample 1}

\textbf{Related question 1:} Let \( z = 2 + \sqrt{2} - (3 + 3 \sqrt{2})i \), and let \( c = 2 - 3i \). Let \( w \) be the result when \( z \) is rotated around \( c \) by \( \frac{\pi}{4} \) counter-clockwise.
Find \( w \).

\textbf{Problem type:} Complex number operation.

\textbf{Key words and relevant words:} complex number, rotation, counter-clockwise, center of rotation, angle.

\textbf{Problem solving strategy:}  The transformation involves translating the system so that the center of rotation aligns with the origin, applying a complex exponential rotation to achieve the desired angular displacement, and then translating back to the original coordinate system. This process ensures that the rotated point maintains its relative position to the center while undergoing the specified rotation. The final result is expressed in terms of its real and imaginary components to provide a complete representation in the complex plane.

\vspace{2em}
\textbf{Sample 2}

\textbf{Related question 2:} \[
\text{The function } f(z) = \frac{(-1 + i \sqrt{3}) z + (-2 \sqrt{3} - 18i)}{2} \text{ represents a rotation around some complex number } c.
\]
 \text{ Find } c.
 
 \textbf{Problem type:} Complex number operation.
 
\textbf{Key words and relevant words:}  function, rotation, complex number, transformation.

\textbf{Problem solving strategy:} The transformation follows a structured process, beginning with a translation to align the rotation center with the origin, followed by the application of a complex linear mapping that encodes both rotation and translation. The fixed point of this transformation, obtained by solving \( f(c) = c \), determines the invariant center around which the system rotates. By decomposing this result into its real and imaginary components, a complete representation of the transformation in the complex plane is achieved.

\subsection{Meta Prompt}
\label{meta}

\begin{figure}[H]
    \centering
    \includegraphics[width=0.85\textwidth]{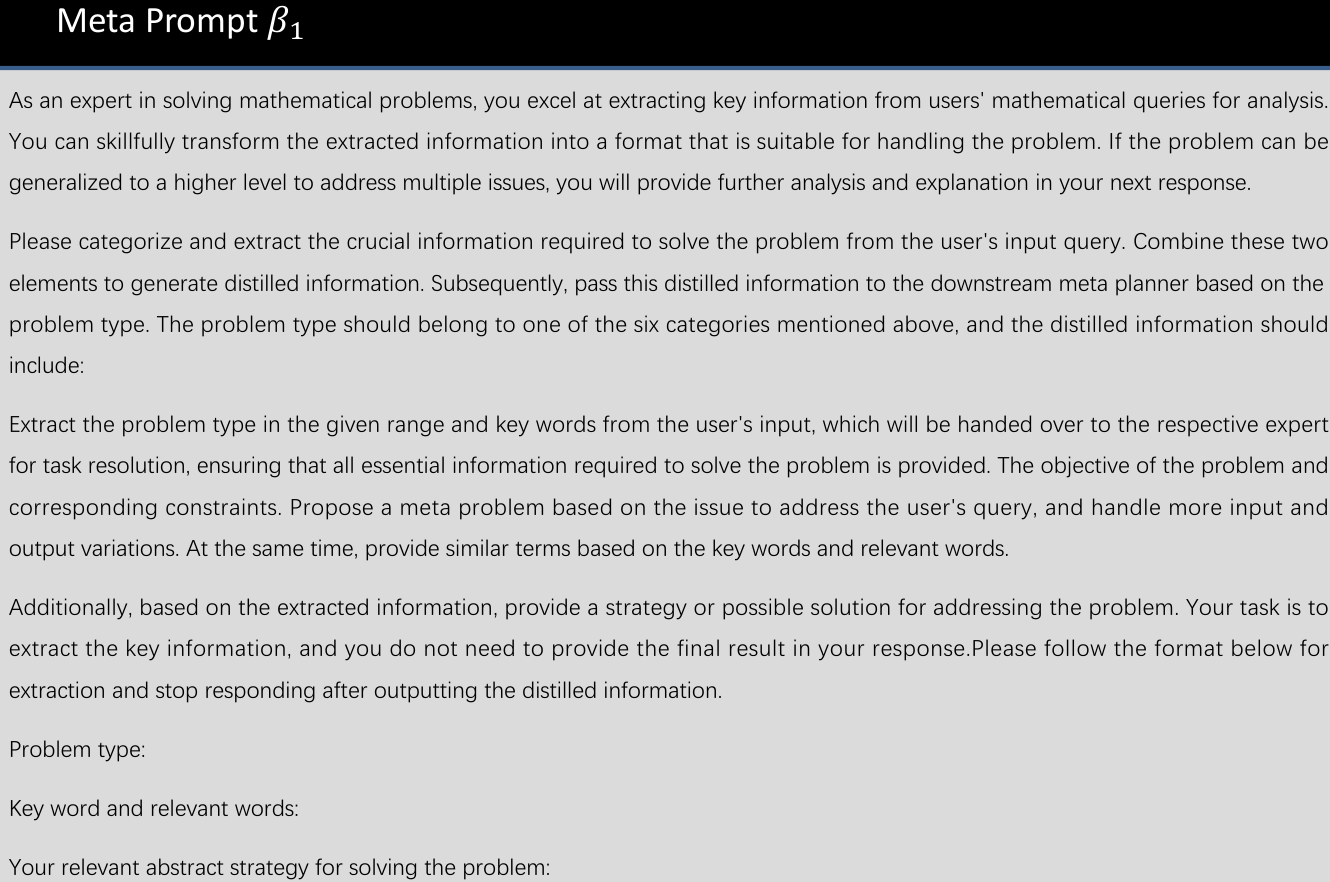}
    \vspace{-0.15in}
    \caption{\small{Meta Prompt $\beta_1$}}
    \label{beta_1}
\end{figure}

\begin{figure}[H]
    \centering
    \includegraphics[width=0.85\textwidth]{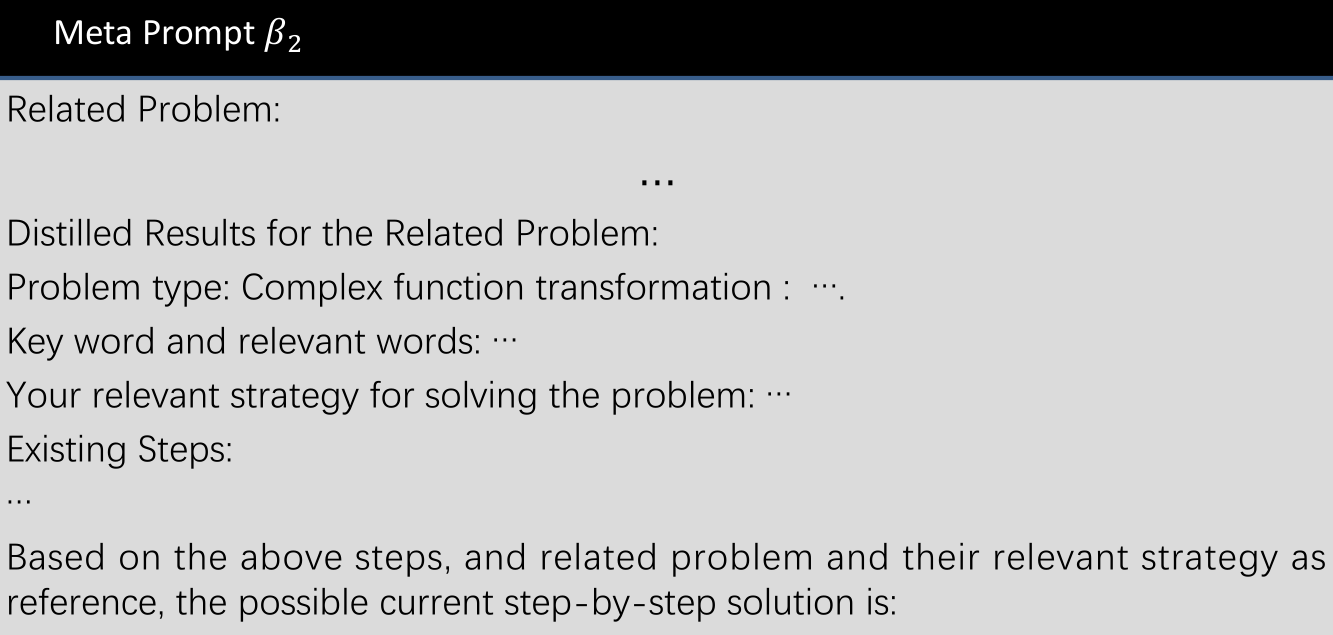}
    \vspace{-0.15in}
    \caption{\small{Meta Prompt $\beta_2$}}
    \label{beta_2}
\end{figure}

%


\begin{figure}[H]
    \centering
    \includegraphics[width=0.85\textwidth]{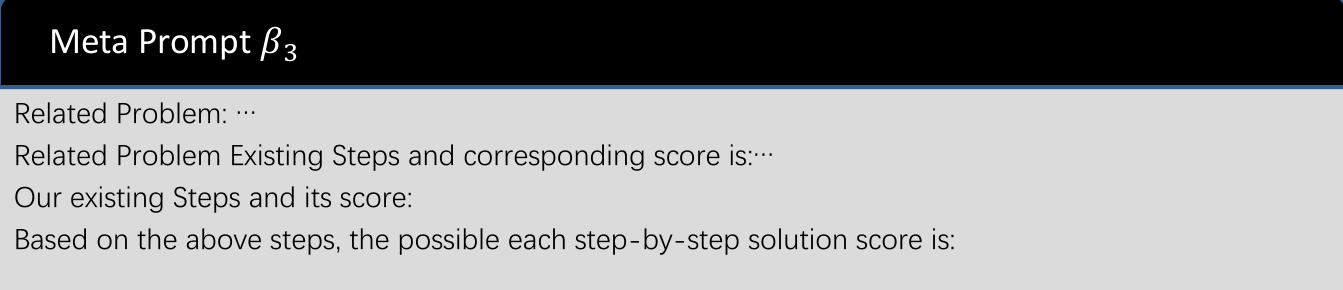}
    \vspace{-0.15in}
    \caption{\small{Meta Prompt $\beta_3$}}
    \label{beta_3}
\end{figure}

Figures \ref{beta_1}, \ref{beta_2}, and \ref{beta_3} illustrate three distinct meta prompts, each designed to assist the large model in a specific task: extracting conceptual units, enhancing DLR reasoning, and refining fine-grained details.

\subsection{Preliminary Filtering \& Refined Selection example}

\begin{tcolorbox}[title=Initial Question]
\textit{"A 90° rotation around the origin transforms the complex number (7+2i). What is the result?"}
\end{tcolorbox}

\vspace{1em}

\textbf{Step 1: Preliminary Filtering (BM25)}

\begin{itemize}
    \item \textbf{Input:} Query = (problem text, type="complex number rotation").
    
    \item \textbf{Candidate Questions Retained (Top 3 by BM25):}
    \begin{enumerate}
        \item "Rotate (3+4i) by 180° around the origin."
        \item "Find the result of rotating (1+i) by 45° counter-clockwise."
        \item "Let (z=2+3i). Rotate (z) by 90° around (1+i)."
    \end{enumerate}
    
    \item \textbf{Rationale:} BM25 prioritizes questions with overlapping keywords (e.g., "rotate", "complex number") and matching problem types.
\end{itemize}

\vspace{1em}

\textbf{Step 2: Refined Selection (SentenceBERT)}

\begin{itemize}
    \item \textbf{Conceptual Unit of Initial Question:}
    \begin{itemize}
        \item ($T_{\text{key}}$): ["origin", "counter-clockwise", "complex number"]
        \item ($T_{\text{strategy}}$): "Apply rotation matrix to complex coordinates."
    \end{itemize}

    \item \textbf{Scores ($S_{\text{ref},i}$):}
    \begin{itemize}
        \item Candidate 1: 0.82 (high, shares "origin" and strategy).
        \item Candidate 2: 0.45 (low, angle differs).
        \item Candidate 3: 0.12 (discarded, rotation center mismatch).
    \end{itemize}

    \item \textbf{Output (DLR Reference Set):} Includes Candidate 1’s solution steps (e.g., "Multiply by \(e^{i\pi}\)" for 180° rotation).
\end{itemize}

\end{document}